\def\year{2018}\relax
\documentclass[letterpaper]{article} %DO NOT CHANGE THIS
\usepackage{graphicx,booktabs,array}
\usepackage{scalerel}

\usepackage{aaai18}  %Required
\usepackage{times}  %Required
\usepackage{helvet,dsfont}  %Required
\usepackage{courier}  %Required
\usepackage{url}  %Required
\usepackage{graphicx}  %Required
\usepackage[utf8]{inputenc} % allow utf-8 input
\usepackage[T1]{fontenc}    % use 8-bit T1 fonts
\usepackage{amsfonts}       % blackboard math symbols
\usepackage{amsmath,amsthm,amssymb,times}
\usepackage{graphicx}
\usepackage{subcaption}
\usepackage[]{algorithm2e}
\newtheorem{non-theorem}{Theorem}

\newtheorem{definition}{Definition}

\newtheorem{example}{Example}

\SetKwInput{KwInput}{Input}
\SetKwInput{KwOutput}{Output}

\def \y {\mathbf y}
\def \eps {\epsilon}

\def \z {\mathbf z}

\def \LB {{\mathrm{LB}}}
\def \UB {{\mathrm{UB}}}
\def \Gen {{\mathrm{Gen}}}
\def \Ver {{\mathrm{Ver}}}
\def \vars {{\mathrm{vars}}}

\def \a {\mathbf a}
\def \b {\mathbf b}

\def \x {\mathbf x}
\def \y {\mathbf y}

\def \v {\mathbf v}
\def \z {\mathbf z}
\def \w {\mathbf w}
\def \l {\mathbf l}

\def \R {\mathbb{R}}

\def \Z {\mathbb{Z}}
\def \nk {{n_{\scaleto{k\mathstrut}{4pt}}}}
\def \nkk {{n_{\scaleto{k+1\mathstrut}{4pt}}}}
\def \nd {{n_{\scaleto{d\mathstrut}{4pt}}}}
\def \bin {^{\scaleto{(b)\mathstrut}{5pt}}}

\newcommand{\PT}{\tau}
\newcommand{\BNN}{\mbox{\sc BNN}}

\newcommand{\BN}{\mbox{\sc Bn}}
\newcommand{\LF}{\mbox{\sc L}}
\newcommand{\BBLOCK}{\mbox{\sc BinBlk}}
\newcommand{\BBO}{\mbox{\sc BinO}}

\newcommand{\FNN}{\mbox{\sc F}}

\newcommand{\CNF}{\mbox{\sc CNF}}
\newcommand{\BLOCK}{\mbox{\sc Blk}}

\newcommand{\LIN}{\mbox{\sc Lin}}
\newcommand{\BIN}{\mbox{\sc Bin}}
\newcommand{\ARGMAX}{\mbox{\sc argmax}}
\newcommand{\SQ}{\mbox{\sc Sq}}

\newcommand{\BO}{\mbox{\sc O}}
\newcommand{\sign}{\mathrm{sign}}

\newcommand{\BBNN}{\mbox{\sc BNN}}

\newcommand{\EILP}{\mbox{\sc \texttt{ILP}}}
\newcommand{\ESAT}{\mbox{\sc \texttt{SAT}}}
\newcommand{\ECEG}{\mbox{\sc \texttt{CEG}}}

\begin{document}
\title{Verifying Properties of Binarized Deep Neural Networks}
%

%\author{
%Nina Narodytska$^1$\hfill
%Shiva Prasad Kasiviswanathan$^2$\hfill
%Leonid Ryzhyk$^1$\\
%\large\textbf{Mooly Sagiv$^1$\hfill
%Toby Walsh$^3$}\\
%$^1$VMware Research, Palo Alto, USA \ \
%$^2$Amazon, Palo Alto, USA \ \
%$^3$UNSW, Sydney, Australia
%}
%\author{~~~Nina Narodytska \\
%~~VMware Research\\
%~~Palo Alto, USA
%%\\n.narodytska@vmware.com
%\And
%Shiva Kasiviswanathan~~~~~\\Amazon~~~~~\\ ~~~~~Palo Alto, USA~~~~~ \newline
%%\\ kasivisw@gmail.com
%\And
%~~~~~Leonid Ryzhyk  ~~
%Mooly Sagiv  \\
%~~~~~VMware Research\\
%~~~~~Palo Alto, USA
%%\\msagiv@vmware.com
%\And
%Toby Walsh \\
%UNSW and Data61\\
%Sydney, Australia
%%\\Toby.Walsh@data61.csiro.au
%}

\author{~~~~~~~Nina Narodytska \\
~~~~~~~VMware Research\\
~~~~~~~Palo Alto, USA
%\\n.narodytska@vmware.com
\And
~~~~~~~~~~Shiva Kasiviswanathan\\~~~~~~~~~~Amazon\\~~~~~~~~~~Palo Alto, USA\\
%\\ kasivisw@gmail.com
\And
~~~~~~~~~~Leonid Ryzhyk\\
~~~~~~~~~~~~VMware Research\\
~~~~~~~~~~~~Palo Alto, USA
\And
~~Mooly Sagiv  \\
~~~VMware Research\\
~~Palo Alto, USA
%\\msagiv@vmware.com
\And
Toby Walsh~~~~~~~\\
UNSW and Data61~~~~~~~\\
Sydney, Australia~~~~~~~
%\\Toby.Walsh@data61.csiro.au
}

%\author{Nina Narodytska$^1$
%\hfill
%Shiva Kasiviswanathan$^2$
%\hfill
%Leonid Ryzhyk$^1$
%\hfill
%Mooly Sagiv$^1$
%\hfill
%Toby Walsh$^3$ \\
%$^1$VMware Research, Palo Alto, USA \ \
%$^2$Amazon, Palo Alto, USA \ \
%$^3$UNSW and Data61, Sydney, Australia
%}

%\affil[1]{Department of Computer Science, \LaTeX\ University}
%\affil[2]{Department of Mechanical Engineering, \LaTeX\ University}
%\affil[3]{Department of Mechanical Engineering, \LaTeX\ University}
\sloppy
\maketitle
\begin{abstract}
Understanding properties of deep neural networks is an important challenge in deep learning. In this paper, we take a step in this direction by proposing a rigorous way of verifying properties of a popular class of neural networks,  Binarized Neural Networks, using the well-developed means of Boolean satisfiability. Our main contribution is a construction that creates a representation of a binarized neural network as a Boolean formula. Our encoding is the first \emph{exact} Boolean representation of a deep neural network. Using this encoding, we leverage the power of modern SAT solvers along with a proposed counterexample-guided search procedure to verify various  properties of these networks. A particular focus will be on the critical property of robustness to adversarial perturbations. For this property, our experimental results demonstrate that our approach scales to medium-size deep neural networks used in image classification tasks.  To the best of our knowledge, this is the first work on verifying properties of deep neural networks using an exact Boolean encoding of the network.
\end{abstract}

\section{Introduction}
Deep neural networks have become ubiquitous in machine learning with applications ranging from computer vision to speech recognition and natural language processing. Neural networks demonstrate excellent performance in many practical problems, often beating specialized algorithms for these problems, which led to their rapid adoption in industrial applications.
With such a wide adoption, important questions arise regarding our understanding of these neural networks: How robust are these networks to perturbations of inputs? How critical is the choice of one architecture over the other?
Does certain ordering of transformations matter?
Recently, a new line of research on understanding neural networks has emerged that look into a wide range of such questions, from interpretability of neural networks to verifying their properties~\cite{BauZKOT17,Szegedy2013,Bastani2016,HuangKWW17,Katz2017}.

In this work we focus on an important class of neural networks:  Binarized Neural Networks~($\BNN$s)~\cite{BNNNIPS2016}. These networks have a number of important features that are useful in resource constrained environments, like embedded devices or mobile phones. Firstly, these networks are memory efficient, as their parameters are primarily binary. Secondly, they are computationally efficient as all activations are binary, which enables the use of specialized algorithms for fast binary matrix multiplication. These networks have been shown to achieve performance comparable to traditional deep networks that use floating point precision on several standard datasets~\cite{BNNNIPS2016}. Recently, $\BNN$s have been deployed for various embedded applications ranging from image classification~\cite{mcdanel2017embedded} to object detection~\cite{kung2017efficient}.

The goal of this work is to analyze properties of binarized neural networks through the lens of Boolean satisfiability (SAT).  Our main contribution is a procedure for constructing a SAT encoding of a binarized neural network. An important feature of our encoding is that it is {\em exact} and does not rely on any approximations to the network structure. This means that this encoding allows us to investigate properties of $\BNN$s by studying similar properties in the SAT domain, and the mapping of these properties back from the SAT to the neural network domain is exact. To the best of our knowledge, this is the first work on verifying properties of deep neural networks using an exact Boolean encoding of a network. In our construction, we exploit attributes of $\BNN$'s, both {\em functional}, e.g., most parameters of these networks are binary, and {\em structural}, e.g., the modular structure of these networks. While these encodings could be directly handled by modern SAT solvers, we show that one can exploit the structure of these encodings to solve the resulting SAT formulas more efficiently based on the idea of {\em counterexample-guided} search~\cite{Clarke2000,McMillan2005,McMillan2003}.

While our encoding could be used to study many properties of $\BNN$s, in this paper we focus on important properties of network {\em robustness} and {\em equivalence}.
%\newcommand{\resref}[1]{{\bf (\ref{res:#1})}}
%\begin{list}{{\bf (\alph{enumi})}}{\usecounter{enumi}
%\setlength{\leftmargin}{0ex}
%\setlength{\listparindent}{0ex}
%\setlength{\parsep}{1pt}}
\begin{enumerate}
\item Deep neural networks have been shown susceptible to crafted adversarial perturbations which force misclassification of the inputs. Adversarial inputs can be used to subvert fraud detection, malware detection, or mislead autonomous navigation systems~\cite{papernot2016limitations,grosse2016adversarial} and pose a serious security risk (e.g., consider an adversary that can fool an autonomous driving system into not following posted traffic signs). Therefore, {\em  certifiably} verifying robustness of these networks to adversarial perturbation is a question of paramount practical importance.  Using a SAT encoding, we can certifiably establish whether or not a $\BNN$ is robust to adversarial perturbation on a particular image.

\item Problems of verifying whether two networks are equivalent in their operations regularly arise when dealing with network alterations (such as that produced by {\em model reduction} operations~\cite{reagen2017deep}) and input preprocessing. Again, with our SAT encodings, we can check for network equivalence or produce instances where the two networks differ in their operation.
%\end{list}
\end{enumerate}
%{Recently, \cite{Cheng2017} independently proposed an encoding of binarized neural networks into SAT. They used combinational miters as an intermediate representation. They also used a similar counterexample-guided search procedure to scale the encoding along with a number of other optimizations.}

Experimentally, we show that our techniques can verify properties of medium-sized $\BNN$s. In Section~\ref{sec:exp}, we present a set of experiments on the MNIST dataset and its variants. For example, for a $\BNN$  with five fully connected layers, we are able to prove the absence of an adversarial perturbation or find such a perturbation for up to $95\%$ of considered images.

\section{Preliminaries} \label{sec:prop}
\noindent\textbf{Notation.} We denote $[m]=\{1,\ldots,m\}$. Vectors are in column-wise fashion, denoted by boldface letters. For a vector $\v \in \R^m$, we use $(v_1,\dots,v_m)$ to denote its $m$ elements. For $p \geq 1$, we use $\| \v \|_p$ to denote the $L_p$-norm of $\v$. For a Boolean CNF (Conjunctive Normal Form) formula $A$, let $\vars(A)$ denote the set of variables in $A$. We say $A$ is {\em unsatisfiable} if there exists no assignment to the variables in $\vars(A)$ that evaluates the formula to true, otherwise we say $A$ is {\em satisfiable}. Let $A$ and $B$ be Boolean formulas.
We denote $\bar{A}$ the negation of $A$.
We say that $A$ \emph{implies} $B$ ($A \Rightarrow B$) iff $\bar{A} \vee B$ is satisfiable and $A$ is \emph{equivalent} to $B$ ($A \Leftrightarrow B$) iff $A \Rightarrow B$ and $B \Rightarrow A$.

Next, we define the supervised image classification problem that we focus on. We are given a set of training images drawn from an unknown distribution $\nu$ over $\mathbb{Z}^n$, where $n$ here represents the ``size'' of individual images. Each image is associated with a label generated by an unknown function  $\LF: \mathbb{Z}^n \rightarrow [s]$, where $[s]=\{1,\dots,s\}$ is a set of possible labels.
During the training phase, given a labeled training set, the goal is to learn a neural network classifier that can be used for inference: given a new image drawn from the same distribution $\nu$, the classifier should predict its true label. During the inference phase, the network is {\em fixed}. In this work, we study properties of such fixed networks generated post training.

\subsection{Properties of Neural Networks}
%As the encodings we construct provide an exact representation of the network, it can be used to study any property related to the network.
In this section, we define several important properties of neural networks, ranging from robustness to properties related to network structure. As the properties defined in this section are not specific to $\BNN$s, we consider a general feedforward neural network denoted by $\FNN$. Let $\FNN(\x)$ represent the output of $\FNN$ on input $\x$ and $\ell_\x = \LF(\x)$ be the ground truth label of~$\x$. {For example, $\x$ can be an image of a bus and $\ell_\x$ is its ground truth label, i.e. `bus'.}

\subsubsection{Adversarial Network Robustness.}
Robustness is an important property that guards the network against adversarial tampering of its outcome by perturbing the inputs.
In the past few years, modern deep networks have been shown susceptible to crafted adversarial perturbations which force misclassification of the inputs, especially images. Adversarial inputs enable adversaries to subvert the expected system behavior leading to undesired consequences and could pose a security risk when these systems are deployed in the real world.
There are now many techniques for generating adversarial inputs, e.g., see~\cite{Szegedy2013,goodfellow2014explaining,moosavi2015deepfool}.  Therefore, a natural question is to understand
how susceptible a network is to any form of adversarial perturbation~\cite{goodfellow2014explaining,moosavi2015deepfool,Bastani2016}. Informally, a network is robust on an input if small perturbations to the input does not lead to misclassification. More formally,
\begin{definition}[Adversarial Robustness\!\footnote{The definition naturally extends to a collection of inputs.}] \label{def:adv:per}
A feedforward neural network $\FNN$ is $(\epsilon,p)$-robust for an input $\x$ if there does not exist $\PT$, $\|\PT\|_{p} \leq \epsilon$, such that $\FNN(\x  + \PT) \neq \ell_\x$.
\end{definition}
%Since for any fixed vector, the $L_p$-norms monotonically decrease as we increase $p$, if a network is $(\epsilon,p)$-robust then it is also $(\epsilon,p')$-robust for all $p' > p$.
The case of $p=\infty$, which bounds the maximum perturbation applied to each entry in $\x$, is especially interesting and has been considered frequently in the literature on adversarial attacks in deep learning~\cite{goodfellow2014explaining,Bastani2016,Katz2017}.

Another popular definition of robustness comes from the notion of {\em universal adversarial} perturbation as defined by~\cite{Moosavi-Dezfooli2016}.  Intuitively, a universal adversarial perturbation is one that leads to misclassification on most  (say, $\rho$-fraction) of the inputs in a set of images.  Absence of such a perturbation is captured in the following definition of robustness. Let $S$ denote a set of images.
\begin{definition} [Universal Adversarial Robustness] \label{def:univ:adv:per}
A feedforward neural network $\FNN$ is $(\epsilon,p,\rho)$-universally robust for a set of inputs in $S$ if there does not exist $\PT$, $\|\PT\|_{p} \leq \epsilon$, such that $\sum_{\x_i \in S }\mathds{1}_{\FNN(\x_i + \PT) \neq l_{\x_i}}  \geq \rho | S |$.
\end{definition}

\subsubsection{Network Equivalence.} Similar to robustness, a property that is commonly verified is that of {\em equivalence} of networks.
Informally, two networks $\FNN_1$ and $\FNN_2$ are equivalent if these networks generate same outputs on all inputs drawn from the domain.  Let $\mathcal{X}$ denote the domain from which inputs are drawn. In the case of images, $\mathcal{X} = \Z^n$.
\begin{definition} [Network Equivalence]
Two feedforward neural networks $\FNN_1$ and $\FNN_2$ are equivalent if for all $\x \in \mathcal{X}$, $\FNN_1(\x) = \FNN_2(\x)$.
\end{definition}
We describe two common scenarios where such equivalence problem arises in practice.
%\begin{list}{{\bf (\arabic{enumi})}}{\usecounter{enumi}
%\setlength{\leftmargin}{0ex}
%\setlength{\listparindent}{0ex}
%\setlength{\parsep}{1pt}}
\begin{enumerate}
\item Network Alteration: Consider a scenario where a part of the trained  network has been altered to form a new network. This change could arise due to model reduction operations that are commonly performed on deep networks to make them amenable for execution on resource-constrained devices~\cite{reagen2017deep} or they could arise from other sources of noise including adversarial corruption of the network.  The question now is whether the altered network is equivalent to the original network?

\item Augmentation Reordering: Consider a scenario where an input is preprocessed (augmented) before it is supplied to a network. Examples of such preprocessing include geometrical transformations, blurring, etc. Let $f:\mathcal{X}\rightarrow \mathcal{X}$ and $g:\mathcal{X}\rightarrow \mathcal{X}$ be two transformation functions (this extends to more transformation functions too). We want to know how sensitive the network is to the order of applications of $f$ and $g$. For example, given a network $\FNN$, let $\FNN_1$ be the network that applies $f \circ g$ on the input followed by $\FNN$, and $\FNN_2$ be the network that applies $g \circ f$ on the input followed by $\FNN$. The question now is whether $\FNN_1$ is equivalent to $\FNN_2$?
%\end{list}
\end{enumerate}
\section{Binarized Neural Networks} \label{sec:bin}
A binarized neural network ($\BNN$) is a feedforward network where weights and activations are predominantly binary~\cite{BNNNIPS2016}.  It is convenient to describe the structure of a $\BNN$ in terms of composition of blocks of layers rather than individual layers. Each block consists of a collection of linear and non-linear transformations. Blocks are assembled sequentially to form a $\BNN$.

\begin{figure}
\begin{center}
\includegraphics[width=0.45\textwidth]{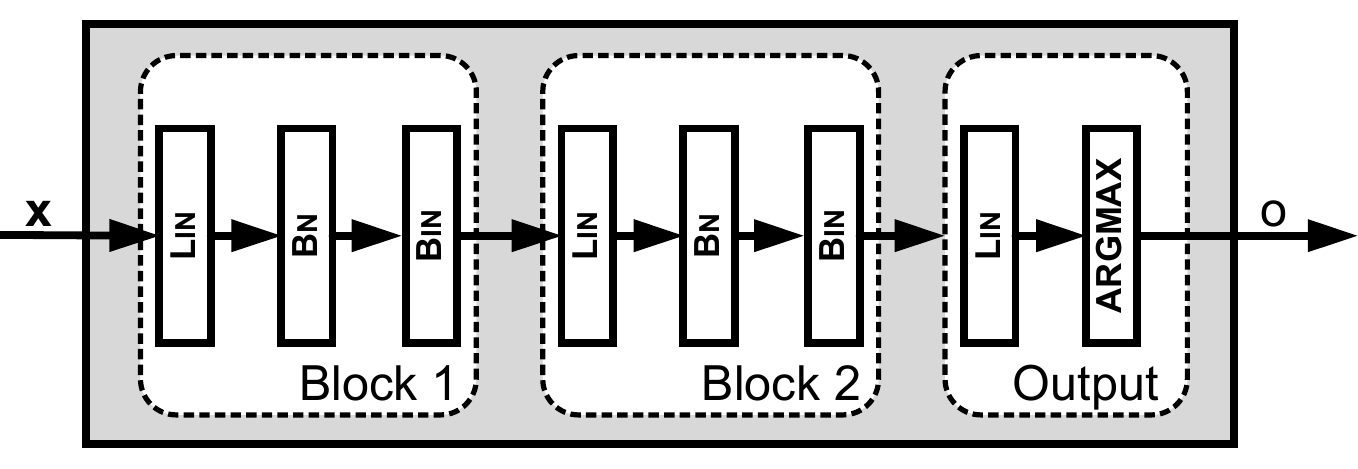}
%\vspace*{-2ex}
\caption{A schematic view of a binarized neural network. The internal blocks also have an additional hard tanh layer.}
\label{fig:lowlevel_bnn_view}
\end{center}
%\vspace*{-5mm}
\end{figure}

\smallskip
\noindent\textbf{Internal Block.} Each internal block (denoted as $\BLOCK$) in a $\BNN$ performs a collection of transformations over a binary input vector and outputs a binary vector. While the input and output of a $\BLOCK$ are binary vectors, the internal layers of $\BLOCK$ can produce real-valued intermediate outputs. A common construction of an internal $\BLOCK$ (taken from~\cite{BNNNIPS2016}) is composed of three main operations:\footnote{In the training phase, there is an additional {\em hard tanh} layer after batch normalization layer that is omitted in the inference phase~\cite{BNNNIPS2016}.}
a linear transformation ($\LIN$), batch normalization ($\BN$), and binarization ($\BIN$). Table~\ref{table:block_bnn} presents the formal definition of these transformations. The first step is a linear (affine) transformation of the input vector. The linear transformation can be based on a fully connected layer or a convolutional layer. The linear  transformation is followed by a scaling which is performed with a batch normalization operation~\cite{Ioffe2015}.  Finally, a binarization is performed using the $\sign$ function to obtain a binary output vector.\!\footnote{A $\BLOCK$ may also contain a max pooling operation to perform  dimensionality reduction, which can be simulated using a convolutional layer with an appropriately chosen stride~\cite{springenberg2015striving}.} Figure~\ref{fig:lowlevel_bnn_view} shows two $\BLOCK$s connected sequentially.

\begin{table*}
\small
\centering
\begin{tabular}{|c|c|}
\hline
\multicolumn{2}{|c|}{Structure of $k$th Internal block, $\BLOCK_k : \{-1,1\}^\nk \rightarrow \{-1,1\}^\nkk$ on input $\x_k \in \{-1,1\}^\nk$} \\
\hline
$\LIN$ & $\y = A_k \x_k + \b_k$ , where $A_k \in \{-1,1\}^{\nkk \times \nk}$ and  $\b_k \in \R^\nkk$\\
$\BN$ & $z_i = \alpha_{k_i} \left ( \frac{y_i - \mu_{k_i}}{\sigma_{k_i}} \right )+\gamma_{k_i}$, where $\y=(y_1,\dots,y_\nkk)$, and $\alpha_{k_i}, \gamma_{k_i}, \mu_{k_i},\sigma_{k_i} \in \R$. Assume $\sigma_{k_i}  > 0$. \\
%$\HT$ & $u_i = \begin{cases}
%1 & \text{if } z_i \geq 1 \\
%-1 & \text{if } z_i \leq -1\\
%z_i & \text{otherwise}
%\end{cases} $\\
$\BIN$ &   $\x_{k+1}= \sign(\z)$ where $\z=(z_1,\dots,z_\nkk) \in \R^\nkk$ and  $\x_{k+1} \in \{-1,1\}^\nkk$\\
\hline
\hline
\multicolumn{2}{|c|}{Structure of Output Block, $\BO : \{-1,1\}^\nd \rightarrow [s]$ on input $\x_d \in \{-1,1\}^\nd$} \\
\hline
$\LIN$ & $\w = A_d \x_d + \b_d$, where $A_d \in \{-1,1\}^{s \times \nd}$ and  $\b_d \in \R^s$\\
$\ARGMAX$ & $o = \mbox{argmax}(\w)$, where $o \in [s]$\\
\hline
\end{tabular}
%\vspace*{-2ex}
\caption{
\small Structure of internal and outputs blocks which stacked together form a binarized neural network. In the training phase, there might be an additional {\em hard tanh} layer after batch normalization. $A_k$ and $b_k$ are parameters of the $\LIN$ layer, whereas $\alpha_{k_i}, \gamma_{k_i}, \mu_{k_i}, \sigma_{k_i}$ are parameters of the $\BN$ layer. The $\mu$'s and $\sigma$'s correspond to mean and standard deviation computed in the training phase. The $\BIN$ layer is parameter free.}\label{table:block_bnn}
%\vspace*{-4mm}
\end{table*}

\smallskip
\noindent\textbf{Output Block.} The output block (denoted as $\BO$) produces the classification decision for a given image. It consists of two layers (see Table~\ref{table:block_bnn}). The first layer applies a linear (affine) transformation that maps its input to a vector of integers, one for each output label class. This is followed by a $\ARGMAX$ layer, which outputs the index of the largest entry in this vector as the predicted label.

\smallskip
\noindent\textbf{Network of Blocks.} $\BNN$ is a deep feedforward network formed by assembling a sequence of internal blocks and an output block. Suppose we have $d-1$ blocks, $\BLOCK_1,\ldots,\BLOCK_{d-1}$ that are placed consecutively, so the output of  a block is an input to the next block in the list. Let $\x_k$ be the input to $\BLOCK_k$ and $\x_{k+1}$ be its output. The input of the first block is the input of the network. We assume that the input of the network is a vector of integers, which is true for the image classification task if images are in the standard RGB format. Note that these integers can be encoded with binary values $\{-1,1\}$ using a standard encoding. Therefore, we keep notations uniform for all layers by assuming that inputs are all binary.
The output of the last layer, $\x_{d} \in \{-1,1\}^\nd$, is passed to the output block $\BO$ to obtain the label.
\begin{definition}[Binarized Neural Network] \label{def:BNN}
A binarized neural network $\BNN:\{-1,1\}^n \rightarrow [s]$ is a feedforward network that is composed of $d$ blocks, $\BLOCK_1,\dots,\BLOCK_{d-1},\BO$. Formally, given an input $\x$,  $$\BNN(\x) = \BO(\BLOCK_{d-1}(\ldots \BLOCK_1(\x) \ldots )).$$
\end{definition}

\section{Encodings of Binarized Networks} \label{sec:enc}
In this section, we consider encodings of $\BNN$s into Boolean formulae. To this end, we encode the two building blocks ($\BLOCK$ and $\BO$) of the network into SAT. The encoding of the $\BNN$ is a conjunction of encodings of its blocks. Let $\BBLOCK_k(\x_k, \x_{k+1})$ be a Boolean function that encodes the $k$th block ($\BLOCK_k$) with an input  $\x_k$ and an output $\x_{k+1}$. Similarly, let $\BBO(\x_d, o)$ be a Boolean function that encodes $\BO$ that takes an input $\x_{d}$ and outputs $o$. The entire $\BNN$ on input $\x$ can be encoded as a Boolean formula (with $\x_1= \x$):
\begin{align*}
\left ( \bigwedge_{k=1}^{d-1} \BBLOCK_k(\x_k, \x_{k+1}) \right ) \wedge \BBO(\x_d,o).
\end{align*}
We consider several encodings of $\BBLOCK$ and $\BBO$ starting with a simple MILP encoding, which is refined to get an ILP encoding, which is further refined to the final SAT encoding.
\subsection{Mixed Integer Linear Program Encoding}
We start with a Mixed Integer Linear Programming (MILP) encoding of $\BLOCK$ and $\BO$ blocks. Our MILP encoding has the same flavor as other encodings in the literature for non-binarized networks, e.g.,~\cite{Bastani2016,Pulina2012} (see Section~\ref{sec:related}).

\smallskip
\noindent\textbf{Encoding of $\BLOCK_k$.} We encode each layer in $\BLOCK_k$ to MILP separately. Let $\a_i$ be the $i$th row of the matrix $A_k$.
Let $\x_k \in \{-1,1\}^\nk$ denote the input to $\BLOCK_k$.

\smallskip
\noindent\textbf{Linear Transformation.} The first layer we consider is the linear layer ($\LIN$). In the following, for simplicity, we focus on the linear transformation applied by a fully connected layer but note that a similar encoding can also be done for a convolution layer (as convolutions are also linear operations). The linear constraint in this case is simply,
\begin{align}
y_i &= \langle \a_i, \x_k \rangle + b_i, i=1,\ldots,\nkk, \label{eq:linear}
\end{align}
where $\y=(y_1,\dots,y_\nkk)$ is a vector in $\R^\nkk$ because the bias part, $\b$, is a vector of real values.

\smallskip
\noindent\textbf{Batch Normalization.}  The second layer is the batch normalization layer ($\BN$) that takes the output of the linear layer as an input. By definition, we have
\begin{align}
z_i = \alpha_{k_i} \left ( \frac{y_i - \mu_{k_i}}{\sigma_{k_i}} \right )+\gamma_{k_i}, \;\; i=1,\ldots,\nkk,  \text{ or } \nonumber \\
\sigma_{k_i} z_i = {\alpha_{k_i}} y_i -\alpha_{k_i} \mu_{k_i} +\sigma_{k_i} \gamma_{k_i},\;\; i=1,\ldots,\nkk \label{eq:bn}.
\end{align}
The above equation is a linear constraint.

\smallskip
\noindent \textbf{Binarization.} For the $\BIN$ operation, which implements a $\sign$ function, we need to deal with conditional constraints.
\begin{align}
z_i \geq 0 & \Rightarrow v_i = 1,\;\; i = 1,\dots,\nkk, \label{eq:sign_plus} \\
z_i < 0 & \Rightarrow v_i = -1,\;\; i = 1,\dots,\nkk \label{eq:sign_neg}.
\end{align}
Since these constraints are implication constraints, we can use a standard trick with the big-$M$ formulation to encode them~\cite{Griva2009}. Define, $\x_{k+1}=(v_1,\dots,v_\nkk)$.

\begin{example}\label{exm:running_example_milp_block}
Consider an internal block with two inputs and one output. Suppose we have the following parameters: $A_k = [1, -1]$, $b_k = [-0.5]$, $\alpha_k = [0.12]$, $\mu_k =  [-0.1]$, $\sigma_k = [2]$ and  $\gamma_k = [0.1]$.
First, we apply the linear transformation: $$y_1 = x_{k_1} - x_{k_2}  - 0.5.$$
Second, we apply  batch normalization:
$$2 z_1 =  0.12y_1 + 0.12 \times 0.1 + 2 \times 0.1.$$
%\begin{align*}
%y_1 &= x_{k_1} - x_{k_2}  - 0.5,\\
%2 z_1 &=  0.12y_1 + 0.12 \times 0.1 + 2 \times 0.1.
%\end{align*}
Finally, we apply binarization. We get $z_1 \geq 0  \Rightarrow v_1 = 1,$  and $ z_1 < 0  \Rightarrow v_1 = -1$.
%\begin{align*}
%z_1 &\geq 0  \Rightarrow v_1 = 1, &\  z_1 &< 0  \Rightarrow v_1 = -1.
%\end{align*}
In this case, $\x_{k+1}$ is just $v_1$.
\end{example}
\smallskip
\noindent\textbf{Encoding of $\BO$.} For the $\BO$ block, encoding of the linear layer is straightforward as in the $\BLOCK$ case:
\begin{align*}
w_i &= \langle \a_i, \x_d \rangle + b_i, \;\; i=1,\ldots,s,
\end{align*}
where $\a_i$ now represents the $i$th column in $A_d$ and $\w=(w_1,\dots,w_s)$. To encode $\ARGMAX$, we need to encode an ordering relation between $w_i$'s. We assume that there is a total order over $w_i$'s for simplicity. % For the encoding,
We introduce a set of Boolean variables $d_{ij}$'s such that
\begin{align*}
w_i \geq w_j & \Leftrightarrow d_{ij} =1, \;\; i,j=1,\ldots,s. %\\
%w_i < w_j & \Rightarrow d_{ij} =0, \;\; i,j=1,\ldots,s.
\end{align*}
Finally, we introduce an output variable $o$ and use more implication constraints.
\begin{align}
\sum_{j=1}^s d_{ij} = s \Rightarrow o = j, \;\; i=1,\ldots,s \label{eq:milp:argmax_bs_geq}.
\end{align}

\begin{example}\label{exm:running_example_milp_output}
Consider an output block with two inputs and two outputs. %, i.e., we have a binary classification problem.
Suppose we have the following parameters for this block $A_d = [1, -1; -1, 1]$ and $b = [-0.5, 0.2]$.
First, we perform encoding of the linear transformation. We get constraints
\begin{align}
w_1 = x_{d_1} - x_{d_2}  - 0.5, \nonumber \\   %\\ %\label{eq:linear_output_example_milp_1}\\
w_2 = -x_{d_1} + x_{d_2}  + 0.2. \nonumber %\label{eq:linear_output_example_milp_2}
\end{align}
As we have only two outputs, we introduce four Boolean variables $d_{ij}$, $i,j = 1,2$.
Note that $d_{11}$ and $d_{22}$  are always true. So, we only need to consider non-diagonal variables. Hence,
we have the following constraints: $w_1 \geq w_2 \Leftrightarrow d_{12} =1$ and
$w_2 < w_1  \Leftrightarrow d_{21} =1$.

Finally, we compute the output $o$ of the neural network as:
$d_{11} + d_{12} = 2 \Rightarrow o = 1$  and
$d_{21} + d_{22} = 2 \Rightarrow o = 2$.
\end{example}
\subsection{ILP Encoding}
In this section,  we show how we can get rid of real-valued variables in the MILP encoding to get a pure ILP encoding which is smaller compared to the MILP encoding.

\smallskip
\noindent\textbf{Encoding of $\BLOCK_k$.}  As the input and the output of $\BLOCK_k$ are integer (binary) values, both $\z$ and $\y$ are functional variables of $\x_k$. Hence, we can substitute them in~\eqref{eq:sign_plus} and in~\eqref{eq:sign_neg} based on~\eqref{eq:linear} and~\eqref{eq:bn} respectively. We get that
\begin{align*}
\frac{\alpha_{k_i}}{\sigma_{k_i}} (\langle \a_i, \x_k \rangle + b_i) - \frac{\alpha_{k_i}}{\sigma_{k_i}}\mu_{k_i} + \gamma_{k_i}  \geq 0 &\Rightarrow v_i = 1.
\end{align*}
Assume $\alpha_{k_i} > 0$, then this translates to
\begin{align*}
\langle \a_i, \x_k \rangle  \geq -\frac{\sigma_{k_i}}{\alpha_{k_i}} \gamma_{k_i}  + \mu_{k_i}  -  b_i & \Rightarrow  x'_i = 1.
\end{align*}
Consider, the left part of above equation. Now, we notice that the left-hand of the equation ($\langle \a_i, \x_k \rangle$) has an integer value, as $\a_i$ and $\x_k$ are binary vectors. Hence, the right-hand side, which is a real value can be rounded safely.  Define
$C_i = \lceil   -\frac{\sigma_{k_i}}{\alpha_{k_i}} \gamma_{k_i}  + \mu_{k_i}  -  b_i \rceil$. Then we can use the following implication constraint to encode~\eqref{eq:sign_plus} and~\eqref{eq:sign_neg}.
\begin{align}
\langle \a_i, \x_k \rangle  \geq C_i & \Rightarrow  v_i = 1,\;\; i=1, \ldots,\nkk, \label{eq:ilpenc_1} \\
\langle \a_i, \x_k \rangle  < C_i  & \Rightarrow  v_i = -1,\;\; i=1, \ldots,\nkk. \label{eq:ilpenc_2}
\end{align}
If $\alpha_{k_i} < 0$, then the idea is again similar, but now we define $C_i = \lfloor -\frac{\sigma_{k_i}}{\alpha_{k_i}} \gamma_{k_i}  + \mu_{k_i} -  b_i  \rfloor$, and use this value of $C_i$ in~\eqref{eq:ilpenc_1} and~\eqref{eq:ilpenc_2}. If $\alpha_{k_i} = 0$, then we can eliminate these constraints and set $v_i$'s based on the sign of~$\gamma_{k_i}$'s.
\begin{example}\label{exm:running_example_ilp_block}
Consider the internal block from Example~\ref{exm:running_example_milp_block}. Following our transformation, we get the following constraints:
\begin{align*}
x_{k_1} - x_{k_2}  \geq \lceil -2 \times 0.1/0.12 - 0.1 + 0.5 \rceil \Rightarrow v_1 = 1, \\
x_{k_1} - x_{k_2}  <  \lceil -2 \times  0.1/0.12 - 0.1 + 0.5 \rceil \Rightarrow v_1 = -1.  %\\
\end{align*}
\end{example}

\smallskip
\noindent\textbf{Encoding of $\BO$.} Next, we consider the output block. As with the MILP encoding, we introduce the Boolean variables $d_{ij}$ but avoid using the intermediate variables $w_i$'s. This translates to:
\begin{align*}
\langle \a_i, \x_d \rangle + b_i \geq \langle \a_j, \x_d \rangle + b_j & \Leftrightarrow  d_{ij} = 1,\;\; i,j =1, \ldots, s.
\end{align*}
where $\a_i$ and $\a_j$ denote the $i$th and $j$th rows in the matrix $A_d$. The above constraints can be reexpressed as:
\begin{align}
\langle \a_i - \a_j, \x_d \rangle \geq  \lceil b_j - b_i \rceil & \Leftrightarrow  d_{ij} = 1, i,j =1,\dots s. \label{eq:ilp:argmax_1_1}
\end{align}
The rounding is sound as $A_d$ and $\x_d$ have only integer (binary) entries. Constraints from~\eqref{eq:milp:argmax_bs_geq} can be reused as all variables there are integers.

\begin{example}\label{exm:running_example_ilp_output}
Consider the output block from Example~\ref{exm:running_example_milp_output} and constraints for $d_{12}$ (encoding for $d_{21}$ is similar).
We follow the transformation above to get:
$$x_{d_1} - x_{d_2}  - 0.5 \geq  -x_{d_1} + x_{d_2}  + 0.2   \Leftrightarrow d_{12} =1,$$
\noindent which can be reexpressed as:
$$x_{d_1} - x_{d_2}   \geq  \lceil0.7/2\rceil   \Leftrightarrow d_{12} =1.$$
\end{example}

\subsection{SAT (CNF) Encoding}
Our next step will be to go from the ILP to a pure SAT encoding. A trivial way to encode the $\BLOCK$ and $\BO$ blocks into CNF is to directly translate their ILP encoding defined above into SAT, but this is inefficient as the resulting encoding will be very large and the SAT formula will blow up even with logarithmic encoding of integers.  In this section, we will further exploit properties of binarized neural networks to construct a more compact SAT encoding. We first recall the definition of {\em sequential counters} from~\cite{Sinz2005,Sinz2005TR} that are used to encode cardinality constraints.

\smallskip
\noindent\textbf{Sequential Counters.} Consider a cardinality constraint: $\sum_{i=1}^m l_i \geq C$, where $l_i \in \{0,1\}$ is a Boolean variable and $C$ is a constant. This can be compiled into $\CNF$ using sequential counters, that we denote as $\SQ(\l,C)$ with $\l=(l_1,\dots,l_m)$.
{Next we recall sequential counters encoding.}
%, to compile it into $\CNF$. %(see Appendix~\ref{app:seq_counters}).
%An auxiliary variable $r_{(m,C)}$ encodes whether $\sum_{i=1}^m l_i$ is greater or equal to~$C$.

Consider a cardinality constraint: $\sum_{i=1}^m l_i \geq C$, where $l_i \in \{0,1\}$ is a Boolean variable and $C$ is a constant.
Then the sequential counter encoding, that we denote as $\SQ(\l,C)$ with $\l=(l_1,\dots,l_m)$, can be written as the following Boolean formula:
\begin{align}
  %\;\; \wedge \nonumber\\
 ({l}_1 \Leftrightarrow r_{(1,1)}) \wedge (\bar{r}_{(1,j)},\;\; j \in \{2,\dots,C\}) \;\; \wedge \nonumber\\
 r_{(i,1)} \Leftrightarrow {l}_i \vee r_{(i-1,1)} \;\; \wedge \nonumber\\
 r_{(i,j)} \Leftrightarrow {l}_i \wedge r_{(i-1,j-1)} \vee r_{(i-1,j)}, j \in \{2,\dots,C\}, \;\; %\wedge \nonumber\\
 %(w_i \Leftrightarrow  l_i \wedge r_{(i-1,C)}) \;\; % \wedge %\nonumber\\
 %(w_m \Leftrightarrow {l}_{m} \wedge {r}_{(m-1,C)}) 
 \label{eqn:SQ}
\end{align}
%
%
%\begin{align}
% \bar{r}_{1,j},\;\; j \in \{2,\dots,C\} \;\; \wedge \nonumber\\
% (\bar{l}_1 \vee r_{(1,1)}) \;\; \wedge %\nonumber\\
% (\bar{l}_i \vee r_{(i,1)}) \;\; \wedge %\nonumber\\
% (\bar{r}_{i-1,1} \vee r_{(i,1)}) \;\; \wedge \nonumber\\
% \bar{l}_{i} \vee \bar{r}_{(i-1,j-1)} \vee r_{(i,j)},\;\; j \in \{2,\dots,C\} \;\; \wedge \nonumber\\
% \bar{r}_{i-1,j} \vee r_{(i,j)},\;\; j \in \{2,\dots,C\} \;\; \wedge \nonumber\\
% (\bar{l}_{i} \vee \bar{r}_{(i-1,C)}) \;\; \wedge % \nonumber\\
% (\bar{l}_{m} \vee \bar{r}_{(m-1,C)}) \label{eqn:SQ},
%\end{align}
where $i \in \{2,\dots,m\}$.
Essentially, this encoding computes the cumulative sums $\sum_{i=1}^p l_i$ and the computation is performed in unary, i.e., the Boolean variable $r_{(j,p)}$
is true iff $\sum_{i=1}^j l_i \geq p$. 
%The overflow bit $v_m$ is set to true if the partial sum $\sum_{i=1}^m l_i$ is greater than  $C$. 
In particular, $r_{(m,C)}$ encodes whether $\sum_{i=1}^m l_i$ is greater than or equal to $C$.

\smallskip
\noindent\textbf{Encoding of $\BLOCK_k$.}
Looking at the constraint in~\eqref{eq:ilpenc_1} (constraint in~\eqref{eq:ilpenc_2} can be handed similarly), we note that the only type of constraint that we need to encode into SAT is a constraint of the form
$\langle \a_i,\x_k  \rangle  \geq C_i \Rightarrow v_i = 1$.  Moreover, we recall that all values in $\a_i$'s are binary and $\x_k$ is a binary vector.
Let us consider the left part of~\eqref{eq:ilpenc_1} and rewrite it using the summation operation:
$\sum_{j=1}^\nk a_{ij} x_{k_j}  \geq C_i$,
where $a_{ij}$ is the $(i,j)$th entry in $A_k$ and $\x_{k_j}$ is the $j$ entry in $\x_k$. Next, we perform a variable replacement where we replace $x_{k_j} \in \{-1,1\}$ with a Boolean variable $x_{k_j}\bin \in \{0,1\}$, with $x_{k_j} = 2x\bin_{k_j} - 1$. Then, we can rewrite the summation as,
\begin{align}
\sum_{j=1}^\nk a_{ij}(2x\bin_{k_j} - 1)  \geq C_i. \label{eq:sat:start_eq_2}
\end{align}
Denote $\a_i^+ = \{j \mid a_{ij} = 1\}$ and $\a_i^- = \{j \mid a_{ij} = -1\}$. Now~\eqref{eq:sat:start_eq_2} can be rewritten as:
\begin{align}
\sum_{j \in \a_i^+} x\bin_{k_j} -  \sum_{j \in \a_i^-} x\bin_{k_j} \geq \lceil C_i/2 + \sum_{j=1}^\nk a_{ij}/2 \rceil,  \nonumber\\
\sum_{j \in \a_i^+} x\bin_{k_j} -  \sum_{j \in \a_i^-} (1 - \bar{x}\bin_{k_j}) \geq  C'_i, \nonumber\\
\sum_{j \in \a_i^+} x\bin_{k_j} +  \sum_{j \in \a_i^-} \bar{x}\bin_{k_j} \geq C'_i + |\a_i^-| \label{eq:sat:start_eq_full_3},
\end{align}
where $C'_i   = \lceil C_i/2 + \sum_{j=1}^\nk a_{ij}/2 \rceil$.
Let $l_{k_j} = x\bin_{k_j}$ if  $j \in \a_i^+$ and let $l_{k_j} = \bar{x}\bin_{k_j}$ if $j \in \a_i^-$. Let $D_i = C'_i + |\a_i^-|$. Then~\eqref{eq:sat:start_eq_full_3} can be reexpressed as: $\sum_{j=1}^\nk l_{k_j} \geq D_i$,
which is a cardinality constraint. We can rewrite the constraints in~\eqref{eq:ilpenc_1} and \eqref{eq:ilpenc_2} as:
\begin{align}
\sum_{j=1}^\nk l_{k_j} \geq D_i  & \Leftrightarrow v\bin_i = 1,\;\;  i = 1,\ldots,\nkk,  \label{eq:cnf_last}
\end{align}
where $v\bin_i \in \{0,1\}$ is such that $v\bin_i \Rightarrow v_i = 1$ and $\bar{v}\bin_i \Rightarrow v_i = -1$. Notice that~\eqref{eq:cnf_last} is a cardinality constraint conjoined with an equivalence constraint.
Let $\SQ(\l_k,D_i)$ denote the sequential counter encoding of~\eqref{eq:cnf_last} as described above %in Appendix~\ref{app:seq_counters}
with auxiliary variables $r_{i_{(1,1)}},\dots,r_{i_{(\nk,D_i)}}$. Then, $r_{i_{(\nk,D_i)}} \Leftrightarrow v\bin_i$.
%For this, the idea is to encode $\sum_{j=1}^\nk l_j \geq D_i$ using sequential counters (Appendix~\ref{app:seq_counters}). Let $\l=(l_1,\dots,l_\nk)$ and $\SQ(\l,D_i)$ be a sequential counter encoding of a cardinality constraint~\eqref{eq:sat:start_eq_full_4} with auxiliary variables $r_{i_{(1,1)}},\dots,r_{i_{(\nk,D_i)}}$. We can use the sequential counter variable $r_{i_{(\nk,D_i)}}$ to encode the equivalence relation in~\eqref{eq:cnf_last}. Namely, $r_{i_{(\nk,D_i)}} \Leftrightarrow v\bin_i$.

Putting everything together, we define the CNF formula $\BBLOCK_k(\x_k, \x_{k+1})$ for $\BLOCK_k$ as:
$$\bigwedge_{i=1}^\nkk \SQ(\l,D_i) \wedge \bigwedge_{i=1}^\nkk \left (r_{i_{(\nk,D_i)}} \Leftrightarrow v\bin_i \right ),$$
where the construction of the vector $\l_k$ from the input vector $\x_k$ and the construction of the output vector $\x_{k+1}$ from $v\bin_i$'s is as defined above.
\begin{example}\label{exm:running_example_sat_block}
Consider the internal block from Example~\ref{exm:running_example_ilp_block}. Following the above transformation, we can replace $x_{k_1} - x_{k_2}  \geq -1\Leftrightarrow v\bin_1 = 1$
\noindent as follows $2x\bin_{k_1} - 1 - (2x\bin_{k_2} - 1)  \geq -1 \Leftrightarrow v\bin_1 = 1$.

\noindent Then we get
$x\bin_{k_1}  - x\bin_{k_2}   \geq -0.5 \Leftrightarrow v\bin_1 = 1$. This constraint can be reexpressed as
$x\bin_{k_1}  + \bar{x}\bin_{k_2}   \geq  \lceil0.5\rceil \Leftrightarrow v\bin_1 = 1$, which can be encoded using sequential counters.
\end{example}

\smallskip
\noindent\textbf{Encoding of $\BO$.} Consider the constraint in~\eqref{eq:ilp:argmax_1_1}. We observe that the same transformation that we developed for the internal block can be applied here too.  We perform variable substitutions, $x_{d_p} = 2x\bin_{d_p} - 1$, where $x\bin_{d_p}\in\{0,1\}$ and $p \in \{1,\dots,\nd\}$.  Let $E_{ij} = \lceil (b_j - b_i + \sum_{p=1}^\nd a_{ip} - \sum_{p=1}^\nd a_{jp})/2\rceil$. We can reexpress~\eqref{eq:ilp:argmax_1_1} as:
\begin{align*}
\sum_{p \in \a_i^+} x\bin_{d_p} -  \sum_{p \in \a_i^-} x\bin_{d_p} - \sum_{p \in \a_j^+} x\bin_{d_p} +  \sum_{p \in \a_j^-} x\bin_{d_p} \geq  E_{ij},
\end{align*}
where $\a_i^+ = \{p \mid a_{ip}=1\}$ and $\a_i^- = \{p \mid a_{ip}= -1\}$, and similarly $\a_j^+$ and $\a_j^-$. We can simplify this equation where each literal either occurs twice or cancels out as:
%\vspace*{1ex}
\begin{align*}
\sum_{p \in \a_i^+ \cap \a_j^-} x\bin_{d_p} -  \sum_{p \in \a_i^- \cap \a_j^+} x\bin_{d_p}  \geq  \lceil E_{ij}/2 \rceil.
\end{align*}
The rest of the encoding of these constraints is similar to $\BLOCK$ using sequential counters. This finishes the construction of $\BBO(\x_d,o)$.

\begin{example}\label{exm:running_example_sat_output}
Consider the output block from Example~\ref{exm:running_example_ilp_output}. Following the above transformation, we can replace $x_{d_1} - x_{d_2}   \geq  \lceil0.7/2\rceil \Leftrightarrow d_{12} =1$ as follows $2x\bin_{d_1} - 1 - (2x\bin_{d_2} - 1)   \geq  1   \Leftrightarrow d_{12} =1$.
\noindent Then we get $x\bin_{d_1} - x\bin_{d_2}   \geq  0.5   \Leftrightarrow d_{12} =1$. Finally, we rewrite as
 $x\bin_{d_1} + \bar{x}\bin_{d_2}   \geq \lceil  1.5  \rceil  \Leftrightarrow d_{12} =1$ and  can use sequential counters to encode the above expression.
\end{example}

\section{Encoding of Properties} \label{sec:property}
In this section, we use the encoding constructed in the previous section to investigate properties of $\BNN$s defined in Section~\ref{sec:prop}. Note that since our encoding is exact, it allows us to investigate properties of $\BNN$s in the SAT domain.
While our encoding could be used to study any property on these networks, in this paper, we primarily focus on the properties defined in Section~\ref{sec:prop}.
%In this section, we present the encoding for the adversarial robustness property (Definition~\ref{def:adv:per}) and defer others to Appendix~\ref{app:prop}.

\subsubsection{Verifying Adversarial Robustness.} We need to encode the norm restriction on the perturbation and the adversarial condition. Consider an image $\x = (x_1,\dots,x_n)$ and a perturbed image $\x+\PT$, where $\PT=(\PT_1,\dots,\PT_n)$. Our encoding scheme works for any norm constraint that is a linear function of the input, which include the $L_1$- and the $L_\infty$-norms. As discussed in Section~\ref{sec:prop}, the most common norm assumption on $\PT$ is that of
$L_{\infty}$-norm, that we focus on here.

\smallskip
\noindent\textit{Norm Constraint:} For $\| \PT \|_\infty \leq \epsilon$, we need to ensure that $\PT_i \in [-\epsilon,\epsilon]$ for all $i \in [n]$, where $n$ is the size of the images. Note that $\PT_i$ is an integer variable as a valid image has integer values. Therefore, we can use the standard
$\CNF$ conversion from linear constraints over integers into Boolean variables and clauses~\cite{Tamura2009}. Additionally, we add a constraint to ensure that $\x+\PT$ is a valid image. For this, we make a natural assumption that exists a lower bound $\LB$  and an upper bound $\UB$ such that all the entries in a valid image lie within $[\LB,\UB]$. Then we add a constraint to ensure that all the entries in $\x+\PT$ lie within $[\LB,\UB]$.

\smallskip
\noindent\textit{Adversarial Constraint:} We recall that an encoding of a $\BNN$ into SAT contains an integer output variable $o$. The value of $o$ is the predicted label. Hence, we just need to encode that $o \neq \ell_\x$ into a Boolean formula, where $\ell_\x$ is the true label of $\x$. We use $\CNF(\bullet)$ to denote a standard conversion of a constraint over integers into a Boolean formula. Putting these together, checking robustness translates into checking assignments for a formula $\BBNN_{Ad}(\x+\PT, o,\ell_\x)$ defined as:
\begin{align} \label{eqn:advcon}
\BBNN_{Ad}&(\x+\PT, o,\ell_\x) = \CNF(\|\PT\|_{\infty} \leq \epsilon)   \wedge \\  &\bigwedge_{i=1}^{n}\CNF((\x_i+\PT_i) \in [\LB,\UB]) \wedge \nonumber \\
&\BBNN(\x+\PT,o)  
 \wedge  \CNF(o \neq l_X).\nonumber
 %\vspace*{1ex}
\end{align}
We can further simplify this equation by noting that for verifying the adversarial robustness property, we do not need to sort all outputs, but only need to ensure that $l_\x$ is not the top label in the ordering.
%\nina{We recall that we use Boolean variables $d_{ij}$ to sort outputs and to compute the value of the output variable $o$.
%%However, for verifying the adversarial robustness property for one input, we do not need to sort all outputs.
%It is sufficient for us to ensure that the true label $l_X$ is not the top label in the ordering. Therefore, we consider the $l_X$th row of the matrix $d$, namely, $d_{l_X,j}$, $j=1,\ldots,p$ and encode that one of these Boolean variables must take value 0. The variables $d_{ij}$, $i=1,\ldots,s, i\neq l_X$, $j=1,\ldots,s$ and the corresponding constraints are redundant in this case.}

\subsubsection{Verifying Universal Adversarial Robustness.}
This is now a simple extension to~\eqref{eqn:advcon} that verifies universal adversarial robustness.
For each $\x_i \in S$, we create a copy of the adversarial robustness property encoding $\BBNN_{Ad}(\x_i+\PT, o_i,\ell_{\x_i})$, and verify if at least $\rho$-fraction of the inputs are misclassified in $S$. This can be expressed as:
\begin{align*}
\bigwedge_{i=1}^{|S|} \left (\BBNN_{Ad}(\x_i+\PT,o_i,\ell_{\x_i}) \Leftrightarrow q_j \right ) \wedge \\
 \CNF  (\sum_{i=1}^{|S|} q_j \geq \rho |S|  ) .
\end{align*}
%$\bigwedge_{i=1}^{|S|} \left (\BBNN_{Ad}(\x_i+\PT,o_i,\ell_{\x_i}) \Leftrightarrow q_j \right ) \wedge
% \CNF \left (\sum_{i=1}^{|S|} q_j \geq \rho |S| \right )$.
\subsubsection{Verifying Network Equivalence.}
To check the equivalence between two networks $\BNN_1$ and $\BNN_2$ we need to verify that these networks produce the same outputs on all valid inputs $\x$ in the domain. Again assuming that each entry in a valid image lies in the range $[\LB,\UB]$, we can formulate the network equivalence problem as the following decision problem on variables $\x=(x_1,\dots,x_n)$.
\begin{multline}
 \wedge_{i=1}^{n}\CNF(x_i \in [\LB,\UB]) \wedge   \BBNN_1(\x,o_1) \\ \wedge \BBNN_2(\x,o_2)  \wedge (o_1 \neq o_2).  \label{formula:eq}
\end{multline}
If this formula is {\em unsatisfiable} then networks are equivalent. Otherwise, a solution of the formula is a valid witness on which these networks produce different outcomes.

\section{Counterexample-Guided Search Procedure} \label{sec:ceg}
Given the SAT formulas constructed in the previous section, we could directly run a SAT solver on them to verify the desired properties.  However, the resulting encodings could be large with large networks, making them hard to tackle even for state-of-the-art SAT solvers. However, we can take advantage of the modular structure of the network to speed-up the search procedure.  In particular, we use the idea of counterexample-guided search that is
extensively used in formal verification~\cite{Clarke2003,McMillan2005,McMillan2003}.

Observe that the SAT encoding follows the modular structure of the network. Let us illustrate our approach with a simple network consisting of two internal blocks and an output block as in Figure~\ref{fig:lowlevel_bnn_view}.\!\footnote{The approach easily extends to a general network were the partitioning can be applied after each block of layer.} Suppose we want to verify the adversarial robustness property of the network (other properties can be handled similarly). The network can be encoded as a conjunction of two Boolean formulas: $\Gen$ (generator) that encodes the first block of the network, and $\Ver$ (verifier) that encodes the rest of the network. Therefore,
%$\BBNN_{Ad}(\x+\PT, o,\ell_\x) = \Gen(\x+\PT,\y) \wedge \Ver(\y,o,\ell_\x)$, where
%$\Gen(\x+\PT,\y) =  \CNF(\|\PT\|_{\infty} \leq \epsilon) \wedge
%\wedge_{i=1}^{n}\CNF((\x_i+\PT_i) \in [\LB,\UB] \wedge \BBLOCK_1(\x+\PT, \y)$ and
%$\Ver(\y,o,\ell_\x) =  \BBLOCK_2(\y,\z) \wedge \BBO(\z,o) \wedge  \CNF(o \neq l_X)$.
\begin{align*}
& \BBNN_{Ad}(\x+\PT, o,\ell_\x) = \Gen(\x+\PT,\y) \wedge \Ver(\y,\z,o,\ell_\x), \\
& \text{where }  \Gen(\x+\PT,\y) =  \CNF(\|\PT\|_{\infty} \leq \epsilon) \wedge \\
&\wedge_{i=1}^{n}\CNF((\x_i+\PT_i) \in [\LB,\UB] \wedge \BBLOCK_1(\x+\PT, \y); \\ %\mbox{ and}\\
 & \Ver(\y,\z,o,\ell_\x) =  \BBLOCK_2(\y,\z) \wedge \\
 &\BBO(\z,o) \wedge  \CNF(o \neq l_X).
\end{align*}
% We note that
The generator and the verifier only share variables in $\y$  that encode activations shared between Block 1 and Block 2 of the network. The set of variables $\y$ is usually  small compared to all variables in the formula. We exploit this property by using {\em Craig interpolants} to build an efficient search procedure.

\begin{definition} [Craig Interpolants] \label{def:interp}
Let $A$ and $B$ be Boolean formulas %with some shared variables
such that the formula
$A \wedge B$ is unsatisfiable. Then there exists a formula $I$, called interpolant, such that $\vars(I) = \vars(A) \cap \vars(B)$,  $B \wedge I$ is unsatisfiable and $A \Rightarrow I$.
In general, there exist multiple interpolants for the given $A$ and $B$.
\end{definition}
An interpolant can be obtained from a proof of unsatisfiability of $A \wedge B$ produced by a SAT solver~\cite{Pudlak1997}. Note that there could be more than one interpolant $I$, and depending on the technique used for constructing the interpolant, we could get different $I$'s.

Our search procedure first generates a satisfying assignment to variables $\PT$ and $\y$ for the generator formula $\Gen(\x+\PT,\y)$. Let $\hat{\y}$ denote this assignment to $\y$. Then we check if we can extend the assignment $\y=\hat{\y}$ to a satisfying assignment for the verifier formula. If so, we found an adversarial perturbation $\PT$. Otherwise, we generate an interpolant $I$
of $\Gen(\x+\PT,\y) \wedge \Ver(\y=\hat{\y},\z,o,\ell_\x)$ by extracting
an unsatisfiable core of $\Ver(\y=\hat{\y},\z,o,\ell_\x)$~\cite{Book2009}. We use assumptions, which  are assignments of $\hat{\y}$, in the SAT solver to obtain a core. Since none of the satisfying assignments to $I$ can be extended to a valid satisfying assignment of $\BBNN_{Ad}(\x+\PT, o,\ell_\x)$, we block them all in $\Gen$ by redefining: $\Gen := \Gen \wedge \neg I$. Then we repeat the procedure. As we are reducing the solution space in each iteration, this algorithm terminates. If the formula $\Gen(\x+\PT,\y)$ becomes
unsatisfiable, then there is no valid perturbation $\PT$, i.e., the network is $\eps$-robust on  image~$\x$.
\begin{table*}
\scriptsize
\centering
\begin{tabular}{|c||c|c|c||c|c|c||c|c|c||c|c|c|}
\hline
&\multicolumn{9}{|c||}{Solved instances (out of 200)}   & \multicolumn{3}{c|}{Certifiably $\epsilon$-robust} \\
 \cline{2-13}
&\multicolumn{3}{c||}{MNIST } & \multicolumn{3}{c||}{MNIST-rot } & \multicolumn{6}{c|}{MNIST-back-image }  \\
 \cline{2-13}
 &$\ESAT$  &$\EILP$  &$\ECEG$ &$\ESAT$  &$\EILP$  &$\ECEG$ &$\ESAT$  &$\EILP$  &$\ECEG$ & $\ESAT$  &$\EILP$  &$\ECEG$  \\
 \cline{2-13}
 & \#solved (t) &\#solved (t) &\#solved (t) & \#solved (t) &\#solved (t) &\#solved (t) & \#solved (t)&\#solved (t) &\#solved (t) & \#  & \# & \# \\
\hline
$\epsilon =  1$ &
180 (77.3)  &
130 (31.5)  &
171 (34.1)  &
179 (57.4)  &
125 (10.9)  &
197 (13.5)  &
191 (18.3)  &
143 (40.8)  &
191 (12.8)  &
138  &
96  &
138  \\

$\epsilon =  3$ &
187 (77.6)  &
148 (29.0)  &
181 (35.1)  &
193 (61.5)  &
155 (9.3)  &
198 (13.7)  &
107 (43.8)  &
67 (52.7)  &
119 (44.6)  &
20  &
5  &
21  \\

$\epsilon =  5$ &
191 (79.5)  &
165 (29.1)  &
188 (36.3)  &
196(62.7)  &
170(11.3)  &
198(13.7)  &
104 (48.8)  &
70 (53.8)  &
116 (47.4)  &
3  &
--  &
4  \\

\hline
\end{tabular}
%\vspace*{-1ex}
\caption{Results on MNIST, MNIST-rot and MNIST-back-image datasets.  \label{tab:mnist_all}}
%\vspace*{-3ex}
\end{table*}

\section{Related Work} \label{sec:related}
With the wide spread success of deep neural network models, a new line of research on understanding neural networks has emerged that investigate a wide range of these questions, from interpretability of deep neural networks to verifying their properties~\cite{Pulina2010,Bastani2016,HuangKWW17,Katz2017,Cheng2017a}.

\cite{Pulina2010} encode a neural network with sigmoid activation functions as a Boolean formula over linear constraints which has the same spirit as our MILP encoding. They use piecewise linear functions to approximate non-linear activations.  The main issue with their approach is that of scalability as they can only verify properties on small networks, e.g., $20$ hidden neurons. %and $6$ output neurons.
Recently,~\cite{Bastani2016} propose to use an LP solver to verify properties of a neural network. In particular, they show that the robustness to adversarial perturbations  can be encoded as a system of constraints. For computational feasibility, the authors propose an approximation of this constraint system to a linear program. Our approach on the other hand is  an exact encoding that can be used to verify any property of a binarized neural network, not just robustness.
%which is an exact encoding, and can be used to verify any property of a binarized neural network, not just robustness. [Perhaps, they can do other properties as well]
%Our MILP constraint system that we use to encode a binarized neural network uses the idea of Boolean over linear constraints) as in~\cite{Pulina2010} and \cite{Bastani2016}.
{\cite{Cheng2017a}  propose using a MIP solver to verify resilience  properties of a neural network. They proposed a number of  encoding  heuristics to scale the MIP solver and demonstrated the effectiveness and scalability of their approach on several benchmark sets.}

\cite{Katz2017} consider neural networks with ReLU activation functions and show that the Simplex algorithm can be used to deal with them. The important property of this method is that it works with a system of constraints directly rather than its approximation. However, the method is tailored to the ReLU activation function. Scalability is also a concern in this work as  each ReLU introduces a branching choice (an SMT solver is used to handle branching). The authors work with networks with 300 ReLUs in total and 7 inputs. Finally,~\cite{HuangKWW17} perform discretization of real-valued inputs and use a counterexample-guided abstraction refinement procedure to search for adversarial perturbations. The authors assume an existence of inverse functions between layers to perform the abstraction refinement step. Our method also utilizes a counterexample-guided search. However, as we consider blocks of layers rather than individual layers we do not need an abstraction refinement step. Also we use interpolants to guide the search procedure rather than inverse functions.

Finally, to the best of our knowledge, the MILP encoding is the only {\em complete} search procedure for finding adversarial perturbations previously proposed in the literature. In our experiments, as a baseline, we use the ILP encoding (described in Section~\ref{sec:enc}), as it is a more efficient encoding than MILP for $\BNN$s.
Most of the other existing adversarial perturbation approaches are {\em incomplete} search procedures, and are either greedy-based, e.g., the fast gradient method~\cite{goodfellow2014explaining}, or tailored to a particular class of non-linearity, e.g., ReLU~\cite{Katz2017}.

{\cite{Cheng2017} independently proposed an encoding of $\BNN$s into SAT.
They used combinational miters as an intermediate representation. They also used a counterexample-guided like search to scale the encoding along with a number of other optimizations~\footnote{This paper appeared on CoRR while the present paper was under review.}.}
\section{Experimental Evaluation} \label{sec:exp}

%We performed an initial experimental evaluation of the proposed approach.
We used the Torch machine learning framework to train networks on a Titan Pascal X GPU. We trained networks on the MNIST dataset~\cite{lecun1998gradient}, and two modern variants of MNIST: the MNIST-rot  and the MNIST-back-image~\cite{Zhou}. In the MNIST-rot variation of the dataset, the digits were rotated by an angle generated uniformly between 0 and $2 \pi$ radians. In the MNIST-back-image variation of the dataset, a patch from a black-and-white image was used as the background for the digit image.
%All datasets contain $10000$ training images and $50000$ test images.
In our experiments, we focused on the important problem of checking adversarial robustness under the $L_{\infty}$-norm.

%Next we describe the structure of our binarized neural network.
We next describe our $\BNN$ architecture. Our network consists of four internal blocks with each block containing a linear layer ($\LIN$) and a final output block. The linear layer in the first block contains $200$ neurons and the linear layers in other blocks contain $100$ neurons. We use the batch normalization ($\BN$)  and binarization ($\BIN$) layers in each block as detailed in Section~\ref{sec:bin}. Also as mentioned earlier, there is an additional hard tanh layer in each internal block that is only used during training.
%For the  MNIST-rot and MNIST-back-image datasets, we trained a network with five linear layers.
To process the inputs, we add two layers ($\BN$ and $\BIN$) to the $\BNN$, as the first two layers in the network to perform binarization of the gray-scale inputs. This simplifies the network architecture and reduces the search space. This addition decreases the accuracy of the original $\BNN$ network by less than one percent. The accuracy of the resulting network on the MNIST, MNIST-rot, and MNIST-back-image datasets were $95.7\%$, $71\%$ and $70\%$, respectively.
%For example, with the addition of these input processing layers, the accuracy of our network on the MNIST dataset drops from $96.3\%$ to $95.7\%$
%The accuracy of the resulting network on the MNIST-rot and MNIST-back-image datasets were $71\%$ and $70\%$, respectively.

To check for adversarial robustness, for each of the three datasets, we randomly picked $20$ images (from the test set) that were correctly classified by the network for each of the $10$ classes. This resulted in a set of $200$ images for each dataset that we consider for the remaining experiments. To reduce the search space we first focus on important pixels of an image as defined by the notion of {\em saliency map}~\cite{Simonyan2013}. In particular, we first try to perturb the top $50\%$ of highly salient pixels in an image. If we cannot find a valid perturbation that leads to misclassification among this set of pixels then we search again over all pixels of the image.
We experimented with three different maximum perturbation values by varying $\epsilon \in \{1,3,5\}$. The timeout is 300 seconds for each instance to solve.

We compare three methods of searching for adversarial perturbations. The first method is an ILP method where we used the SCIP solver~\cite{MaherFischerGallyetal} on the ILP encoding of the problem (denoted $\EILP$). The second method is a pure SAT method, based on our proposed encoding, where we used the Glucose SAT solver~\cite{Audemard2009} to solve the resulting encoding (denoted $\ESAT$). For the third method, we used the same SAT encoding but for speeding-up the search we utilized the counterexample-guided search procedure described in Section~\ref{sec:ceg}.  We use the core produced by the verifier as an interpolant. % (it might not be the most general core).
We call this method $\ECEG$. On average, our generated SAT formulas contain about $1.4$ million variables and $5$ million clauses.
{MNIST-rot generates the largest encoding with about 7 million clauses on average, whereas, MNIST and MNIST-back have about 5 and 3 million clauses on average, respectively. The size of the encoding depends on network parameters (e.g., the right hand side of cardinality constraints) and pixel values as well}.
The largest instance contains about $3$ million variables and $12$ million clauses.

\begin{figure*}
\centering
\begin{tabular}{ccc}
\begin{subfigure}{0.3\textwidth}\centering\includegraphics[width=0.9\columnwidth]{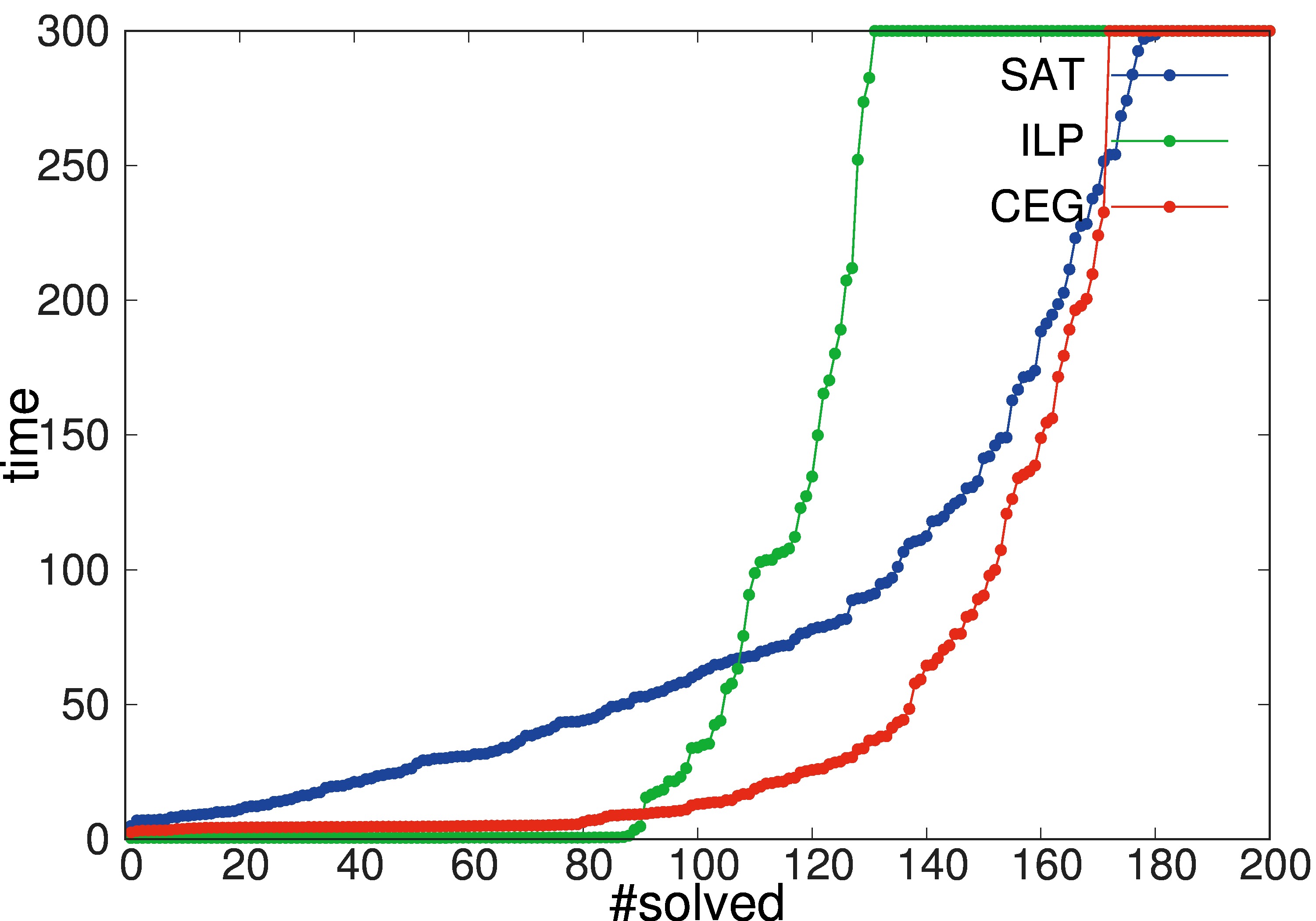}\caption{MNIST, $\epsilon=1$}\end{subfigure}&
\begin{subfigure}{0.3\textwidth}\centering\includegraphics[width=0.9\columnwidth]{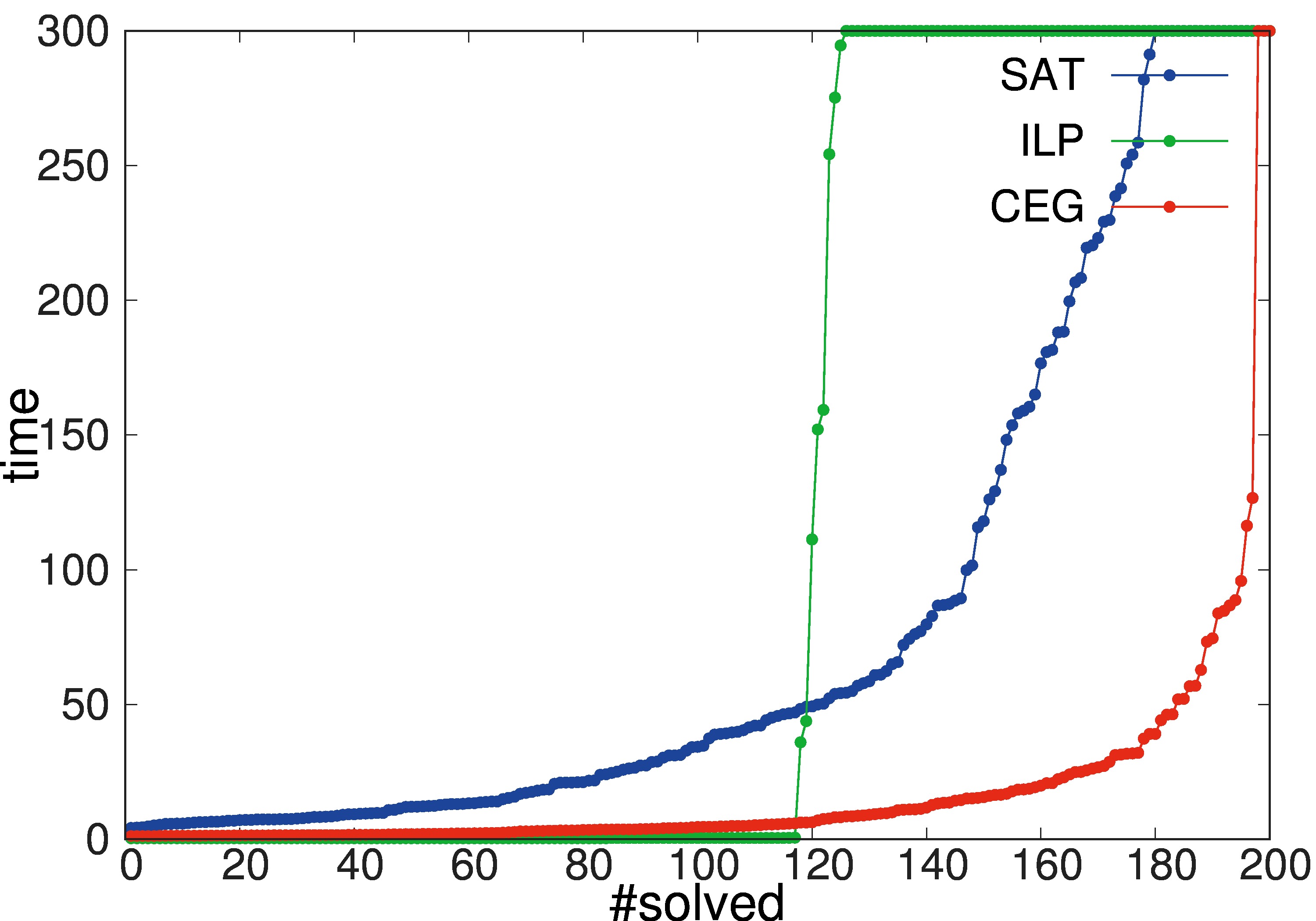}\caption{MNIST-rot, $\epsilon=1$}\end{subfigure}&
\begin{subfigure}{0.3\textwidth}\centering\includegraphics[width=0.9\columnwidth]{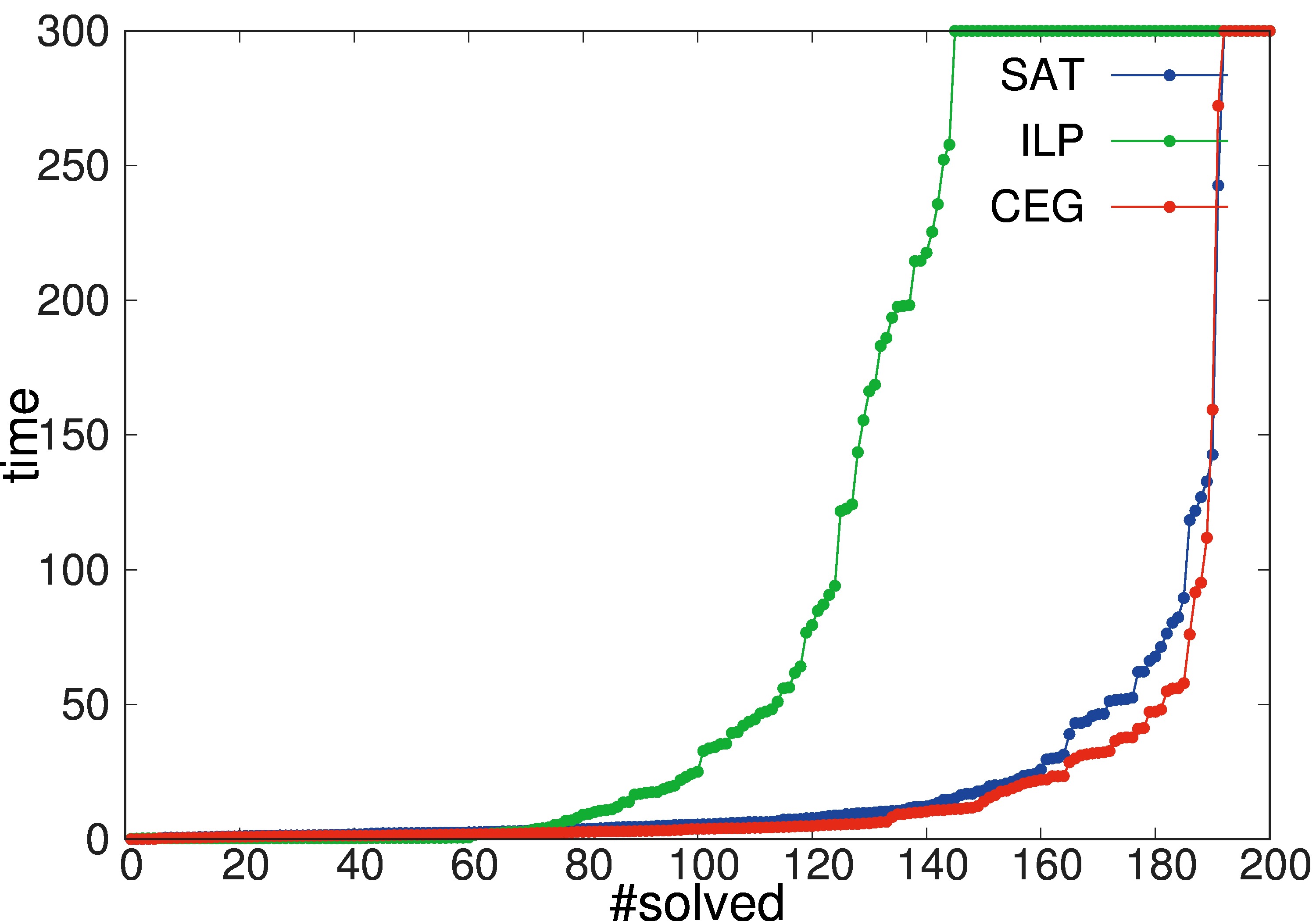}\caption{MNIST-back-image, $\epsilon=1$}\end{subfigure}
\end{tabular}
\caption{Results on MNIST, MNIST-rot and MNIST-back-image datasets with $\epsilon=1$.}
\label{tab:cactus}
\end{figure*}
\begin{figure*}
\centering
\begin{tabular}{cccccc}
\begin{subfigure}{0.07\textwidth}\centering\includegraphics[width=0.9\columnwidth]{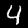}\end{subfigure}&
\begin{subfigure}{0.07\textwidth}\centering\includegraphics[width=0.9\columnwidth]{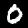}\end{subfigure}&
\begin{subfigure}{0.07\textwidth}\centering\includegraphics[width=0.9\columnwidth]{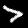}\end{subfigure}&
\begin{subfigure}{0.07\textwidth}\centering\includegraphics[width=0.9\columnwidth]{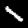}\end{subfigure}&
\begin{subfigure}{0.07\textwidth}\centering\includegraphics[width=0.9\columnwidth]{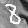}\end{subfigure}&
\begin{subfigure}{0.07\textwidth}\centering\includegraphics[width=0.9\columnwidth]{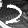}\label{fig:taba}\end{subfigure}\\
\begin{subfigure}{0.07\textwidth}\centering\includegraphics[width=0.9\columnwidth]{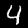}\end{subfigure}&
\begin{subfigure}{0.07\textwidth}\centering\includegraphics[width=0.9\columnwidth]{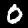}\end{subfigure}&
\begin{subfigure}{0.07\textwidth}\centering\includegraphics[width=0.9\columnwidth]{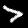}\end{subfigure}&
\begin{subfigure}{0.07\textwidth}\centering\includegraphics[width=0.9\columnwidth]{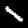}\end{subfigure}&
\begin{subfigure}{0.07\textwidth}\centering\includegraphics[width=0.9\columnwidth]{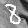}\end{subfigure}&
\begin{subfigure}{0.07\textwidth}\centering\includegraphics[width=0.9\columnwidth]{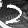}\end{subfigure}
\end{tabular}
\caption{The original (the top row) and successfully perturbed (the bottom row) images.}
\label{tab:images}
\end{figure*}
Table~\ref{tab:mnist_all} presents the results for different datasets. In each column we show the number of instances solved by the corresponding method ({\#solved}) out of the $200$ selected instances and the average time in seconds ({t}) to solve these instances (within the 300 seconds timeout). Solved in this context means that either we determine a valid perturbation leading to misclassification of the image by the network, or concretely establish that there exists no solution which means that the network is $\epsilon$-robust on this image. {Figure~\ref{tab:cactus} compares performance of the algorithms for  $\epsilon=1$ using cactus plots. Cactus plots for other values of $\epsilon$ are similar.}

It is clear that $\ESAT$ and $\ECEG$ methods outperform the $\EILP$ approach. The $\EILP$ method solves up to $30\%$ fewer instances compared to $\ESAT$ or $\ECEG$ across all datasets. This demonstrates the effectiveness of our SAT encoding compared to the ILP encoding. Comparing the $\ESAT$ and $\ECEG$ methods, we see that the results are mixed. On the MNIST-rot and  MNIST-back-image datasets $\ECEG$ solves more instances than $\ESAT$, while on the MNIST dataset we have the opposite situation. We observe that $\ECEG$ is faster compared to $\ESAT$ across all datasets. {From cactus plots, it is clear that $\ECEG$ solves most of the instances faster than $\ESAT$.}
Generally, we noticed that the images in the MNIST-rot dataset were easiest to perturb, whereas images in the MNIST-back-image dataset were the hardest to perturb.

A major advantage of our complete search procedure is that we can {\em certify} $\epsilon$-robustness, in that there exists no adversarial perturbation technique that can fool the network on these images. Such guarantees cannot be provided by incomplete adversarial search algorithms previously considered in the literature~\cite{Szegedy2013,goodfellow2014explaining,moosavi2015deepfool}. In the last three columns of Table~\ref{tab:mnist_all}, we present the number of images in the MNIST-back-image dataset on which the network is certifiably $\epsilon$-robust as found by each of the three methods. Again the SAT-based approaches outperform the ILP approach. With increasing $\epsilon$, the number of images on which the network is $\epsilon$-robust decreases as the adversary can leverage the larger $\epsilon$ value to construct adversarial images. This decrease is also reflected in Table~\ref{tab:mnist_all} (Columns 7--9) in terms of the number of solved instances which decreases from $\epsilon=3$ to $\epsilon=1$, as some of the instances on which a lack of solution can be certified at $\epsilon=1$ by a method cannot always be accomplished at $\epsilon=3$ within the given time limit.

\noindent\textbf{Examples of Perturbed Images.}
Figure~\ref{tab:images} shows examples of original and successfully perturbed images generated by our $\ESAT$ method. We have two images from each dataset in the table. As can be seen from these pictures, the differences are sometime so small that original and perturbed images, that they are indistinguishable for the human eye. These examples illustrate that it is essential that we use a certifiable search procedure to ensure that a network is robust against adversarial inputs.

%In Appendix~\ref{app:per_images}, we present few original and perturbed images found by our approach.

\section{Conclusion}
We proposed an exact Boolean encoding of binarized neural networks that allows us to verify interesting properties of these networks such as robustness and equivalence. We further proposed a counterexample-guided search procedure that leverages the modular structure of the underlying network to speed-up the property verification.  Our experiments demonstrated feasibility of our approach on the MNIST and its variant datasets.  {Our future work will focus on improving the scalability of the proposed approach to enable property verification of even larger neural networks by exploiting the structure of the formula. We see two promising directions here: (1) sharing variables between cardinality constraint encodings, and (2) using counterexample-guided search cuts between multiple layers. Another interesting research direction is to consider more general classes of neural networks that use a fixed number of bits to represent weights. In principle, our approach can handle such networks but the scalability issue is even more challenging there.}

%Another interesting direction is to verify properties of recurrent neural networks that have a number of additional structural properties, e.g., repetitive blocks of layers, that can be exploited during reformulation and search.

%hrough this encoding, we showed that a number of interesting properties of these networks such as robustness and equivalence can be verified. We further proposed a counterexample-guided search procedure that leverages the modular structure of the underlying network to speed-up the property verification. We demonstrated feasibility of our approach on few variants of the MNIST dataset and were able to prove that a number of images in the dataset are robust to adversarial perturbations. Our future work will focus on improving the scalability of the proposed approach to enable property verification of even larger neural networks by exploiting the structure of the formula.
%Another interesting direction is to verify properties of recurrent neural networks that have a number of additional structural properties, e.g., repetitive blocks of layers, that can be exploited during reformulation and search.
\bibliography{lit}
\bibliographystyle{aaai}
\end{document}